\documentclass{article}

\usepackage{graphicx} 
\usepackage{subfigure}

\usepackage{natbib}

\usepackage[breaklinks]{hyperref}

\usepackage[accepted]{icml2012}

\icmltitlerunning{AOSO-LogitBoost}

\usepackage{amsmath}
\usepackage{amssymb}
\DeclareMathOperator*{\argmax}{arg\,max}
\DeclareMathOperator*{\argmin}{arg\,min}
\usepackage{graphicx}
\usepackage{algorithm}
\usepackage{algorithmic}
\usepackage{subfigure}

\newcommand{\RR}{\mathbb{R}}
\newcommand{\xv}{\boldsymbol{x}} 

\newcommand{\fv}{\boldsymbol{f}}
\newcommand{\Fv}{\boldsymbol{F}}
\newcommand{\onev}{\boldsymbol{1}}
\newcommand{\gv}{\boldsymbol{g}}
\newcommand{\Hm}{\boldsymbol{H}}
\newcommand{\pv}{\boldsymbol{p}}

\newcommand{\tv}{\boldsymbol{t}}
\newcommand{\Is}{\mathcal{I}}
\newcommand{\zerov}{\boldsymbol{0}}
\newcommand{\Phat}{\hat{\boldsymbol{P}}}

\newcommand{\ie}{\emph{i.e.}, }

\newtheorem{theo-stzH}{Property}
\newtheorem{lemma-out}{Lemma}

\begin{document}

\twocolumn[
\icmltitle{AOSO-LogitBoost: Adaptive One-Vs-One LogitBoost for Multi-Class Problem}

\icmlauthor{Peng Sun}{sunp08@mails.tsinghua.edu.cn}
\icmladdress{Tsinghua National Laboratory for Information Science and Technology(TNList), Department of Automation, Tsinghua University, Beijing 100084, China}
\icmlauthor{Mark D. Reid}{mark.reid@anu.edu.au}
\icmladdress{Research School of Computer Science, The Australian National University and NICTA, Canberra, Australia}
\icmlauthor{Jie Zhou}{jzhou@tsinghua.edu.cn}
\icmladdress{Tsinghua National Laboratory for Information Science and Technology(TNList), Department of Automation, Tsinghua University, Beijing 100084, China}

\icmlkeywords{boosting, logitboost, multiclass}

\vskip 0.3in
]


\begin{abstract}
This paper presents an improvement to model learning when using multi-class LogitBoost for classification. Motivated by the statistical view, LogitBoost can be seen as additive tree regression. Two important factors in this setting are: 1) coupled classifier output due to a sum-to-zero constraint, and 2) the dense Hessian matrices that arise when computing tree node split gain and node value fittings. In general, this setting is too complicated for a tractable model learning algorithm. However, too aggressive simplification of the setting may lead to degraded performance. For example, the original LogitBoost is outperformed by ABC-LogitBoost due to the latter's more careful treatment of the above two factors.

In this paper we propose techniques to address the two main difficulties of the LogitBoost setting: 1) we adopt a vector tree (\ie each node value is vector) that enforces a sum-to-zero constraint, and 2) we use an adaptive block coordinate descent that exploits the dense Hessian when computing tree split gain and node values. Higher classification accuracy and faster convergence rates are observed for a range of public data sets when compared to both the original and the ABC-LogitBoost implementations.
\end{abstract}

\section{Introduction}
Boosting is successful for both binary and multi-class classification~\citep{fs-boosting,schapire-realboosting}. Among those popular variants, we are particularly focusing on LogitBoost~\cite{Friedman-additive} in this paper. Originally, LogitBoost is motivated by statistical view~\cite{Friedman-additive}, where boosting algorithms consists of three key components: the loss, the function model, and the optimization algorithm. In the case of LogitBoost, these are the Logit loss, the use of additive tree models, and a stage-wise optimization, respectively. There are two important factors in the LogitBoost setting. Firstly, the posterior class probability estimate must be normalised so as to sum to one in order to use the Logit loss. This leads to a coupled classifier output, \ie the sum-to-zero classifier output. Secondly, a dense Hessian matrix arises when deriving the tree node split gain and node value fitting. It is challenging to design a tractable optimization algorithm that fully handles both these factors. Consequently, some simplification and/or approximation is needed. \citet{Friedman-additive} proposes a ``one scalar regression tree for one class'' strategy. This breaks the coupling in the classifier output so that at each boosting iteration the model updating collapses to $K$ independent regression tree fittings, where $K$ denotes the number of classes. In this way, the sum-to-zero constraint is dropped and the Hessian is approximated diagonally.

Unfortunately, Friedman's prescription turns out to have some drawbacks. A later improvement, ABC-LogitBoost, is shown to outperform LogitBoost in terms of both classification accuracy and convergence rate~\citep{Li-ABC,Li-ABCLogit}. This is due to ABC-LogitBoost's careful handling of the above key problems of the LogitBoost setting. At each iteration, the sum-to-zero constraint is enforced so that only $K-1$ scalar trees are fitted for $K-1$ classes. The remaining class -- called the base class -- is selected adaptively per iteration (or every several iterations), hence the acronym ABC (Adaptive Base Class). Also, the Hessian matrix is approximated in a more refined manner than the original LogitBoost when computing the tree split gain and fitting node value.

\begin{figure}
  \centering
  \subfigure[] {
    \includegraphics[width=0.167\textwidth]{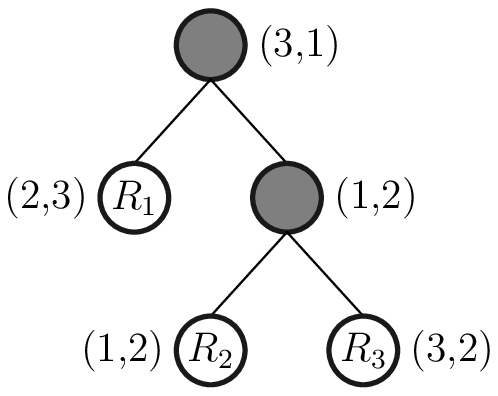}
  }
  \subfigure[] {
    \includegraphics[width=0.25\textwidth]{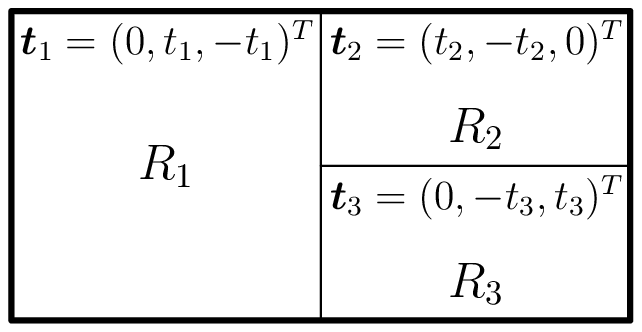}
  }
  \caption{A newly added tree at some boosting iteration for a 3-class problem. \textbf{(a)} A class pair (shown in brackets) is selected for each tree node. For each internal node (filled), the pair is for computing split gain; For terminal nodes (unfilled), it is for node vector updating. \textbf{(b)} The feature space (the outer black box) is partitioned by the tree in (a) into regions $\{R_1,R_2,R_3\}$. On each region only two coordinates are updated based on the corresponding class pair shown in (a).}
  \label{fig-tree-model} 
\end{figure}

In this paper, we propose two novel techniques to address the challenging aspects of the LogitBoost setting. In our approach, one vector tree is added per iteration. We allow a $K$ dimensional sum-to-zero vector to be fitted for each tree node. This permits us to explicitly formulate the computation for both node split gain and node value fitting as a $K$ dimensional constrained quadratic optimization, which arises as a subproblem in the inner loop for split seeking when fitting a new tree. To avoid the difficulty of a dense Hessian, we propose that for \textbf{each} of these subproblems, only two coordinates (\ie two classes or a class pair) are adaptively selected for updating, hence the name AOSO (Adaptive One vS One) LogitBoost. Figure~\ref{fig-tree-model} gives an overview of our approach. In Section~\ref{sec-subset sel} we show that first order and second order approximation of loss reduction can be a good measure for the quality of selected class pair.

Following the above formulation, ABC-LogitBoost, although derived from a somewhat different framework in~\cite{Li-ABCLogit}, can thus be shown to be a special case of AOSO-LogitBoost, but with a less flexible tree model. In Section~\ref{sec-compABC} we compare the differences between the two approaches in detail and provide some intuition for AOSO's improvement over ABC.

The rest of this paper is organised as follows: In Section~\ref{sec-algo} we first formulate the problem setting for LogitBoost and then give the details of our approach. In Section~\ref{sec-comp} we compare our approach to (ABC-)LogitBoost. In Section~\ref{sec-exp}, experimental results in terms of classification errors and convergence rates are reported on a range of public datasets.

\section{The Algorithm}\label{sec-algo}

We begin with the basic setting for the LogitBoost algorithm. For $K$-class classification ($K \geq 2$), consider an $N$ example training set $\{\boldsymbol{x}_i, y_i\}_{i=1}^N$ where $\boldsymbol{x}_i$ denotes a feature value and $y_{i}\in\{1,\ldots,K\}$ denotes a class label. Class probabilities conditioned on $\boldsymbol{x}$, denoted by $\boldsymbol{p} = (p_{1},\ldots,p_{K})^{T}$, are learned from the training set. For a test example with known $\boldsymbol{x}$ and unknown $y$, we predict a class label by using the Bayes rule: $y = \argmax_{k} p_k, k=1,\ldots,K$.

Instead of learning the class probability directly, one learns its ``proxy'' $\boldsymbol{F} = (F_{1},\ldots,F_{K})^{T}$ given by the so-called Logit link function:
\begin{equation}\label{eq-link}
    p_k = \frac  { \exp(F_k) } { \sum_{j=1}^K\exp(F_j) }
\end{equation}
with the constraint $\sum_{k=1}^{K}F_{k} = 0$~\cite{Friedman-additive}. For simplicity and without confusion, we hereafter omit the dependence on $x$ for $\boldsymbol{F}$ and for other related variables.

The $\boldsymbol{F}$ is obtained by minimizing a target function on training data:
\begin{equation}\label{eq-total loss}
    Loss = \sum_{i=1}^{N} L(y_{i}, \boldsymbol{F}_{i}),
\end{equation}
where $\boldsymbol{F}_{i}$ is shorthand for $\boldsymbol{F}(\boldsymbol{x_i})$ and $L(y_{i}, \boldsymbol{F}_{i})$ is the Logit loss for a single training example:
\begin{equation}\label{eq5-log loss}
    L(y_{i}, \boldsymbol{F}_{i}) = -\sum_{k=1}^{K} r_{ik} \log p_{ik},
\end{equation}
where $r_{ik} = 1$ if $y_{i}=k$  and $0$ otherwise. The probability $p_{ik}$ is connected to $F_{ik}$ via (\ref{eq-link}).

To make the optimization of (\ref{eq-total loss}) feasible, a model is needed to describe how $\boldsymbol{F}$ depends on $\boldsymbol{x}$. For example, linear model $\boldsymbol{F} = \boldsymbol{W}^{T}\boldsymbol{x}$ is used in traditional Logit regression, while Generalized Additive Model is adopted in LogitBoost:\vspace{-0.1in}
\begin{equation}\label{eq2-F model}
    \boldsymbol{F(\xv)} = \textstyle\sum_{m=1}^{M}\fv_m(\xv),
\end{equation}
where each $\fv_m(\xv)$, a $K$ dimensional sum-to-zero vector, is learned by greedy stage-wise optimization. That is, at each iteration $\fv_m(\xv)$ is added only based on $\boldsymbol{F} = \sum_{j=1}^{m-1} \boldsymbol{f}_j$. Formally,
\begin{equation}\label{eq-fm}
\begin{split}
    &\fv_m(\xv) = \argmin_{\boldsymbol{f}}
                       \sum_{i=1}^{N} L(y_i,\boldsymbol{F}_i+\boldsymbol{f(\xv_i)}) \\
    &s.t. \sum_k f_k(\xv_i) = 0, \quad i = 1,...,N.
\end{split}
\end{equation}
This procedure repeats $M$ times with initial condition $F=0$. Owing to its iterative nature, we only need to know how to solve (\ref{eq-fm}) in order to implement the optimization.

\subsection{Vector Tree Model}

The $\fv(\xv)$ in (\ref{eq-fm}) is typically represented by $K$ scalar regression trees (\emph{e.g.}, in LogitBoost~\cite{Friedman-additive} or the Real AdaBoost.MH implementation in \cite{Friedman-additive}) or a single vector tree (\emph{e.g.}, the Real AdaBoost.MH implementation in \cite{kegl-boosting}). In this paper, we adopt a single vector tree. We further restrict that it is a binary tree (\ie only binary splits on internal node are allowed) and the split must be vertical to coordinate axis, as in \cite{Friedman-additive} or \cite{Li-ABCLogit}. Formally,
\begin{equation}\label{eq-vector tree}
    \boldsymbol{f}(\xv) = \sum_{j=1}^{J} \boldsymbol{t}_{j} I(\boldsymbol{x} \in R_{j})
\end{equation}
where $\{ R_{j} \}_{j=1}^J$ describes how the feature space is partitioned, while $\boldsymbol{t}_{j} \in \RR^K$ with a sum-to-zero constraint is the node values/vector associated with $R_j$. See Figure~\ref{fig-tree-model} for an example.

\subsection{Tree Building}
Solving (\ref{eq-fm}) with the tree model (\ref{eq-vector tree}) is equivalent to determining the parameters $\{ \boldsymbol{t}_{j}, R_{j} \}_{j=1}^{J}$ at the $m$-th iteration. In this subsection we will show how this problem reduces to solving a collection of convex optimization subproblems for which we can use any numerical method. Following Friedman's LogitBoost settings, here we use Newton descent\footnote{We use Newtown descent as there is evidence in \cite{Li-ABCLogit} that gradient descent, \ie in Friedmans's MART~\cite{Friedman-mart}, leads to decreased classification accuracy.}. Also, we will show how the gradient and Hessian can be computed incrementally.

We begin with some shorthand notation for the \emph{node loss}:
\begin{equation}\label{eq-node loss}
\begin{split}
   &NodeLoss(\boldsymbol{t};\mathcal{I}) = \sum_{i \in \mathcal{I}} L(y_{i}, \boldsymbol{F}_{i}+\boldsymbol{t}) \\
   &t_{1}+\ldots+t_{K} = 0, \quad \tv \in \RR^K
\end{split}
\end{equation}
where $\mathcal{I}$ denotes the index set of the training examples on some either internal or terminal node (\ie those falling into the corresponding region). Minimizing (\ref{eq-node loss}) is the bridge to $\{ \boldsymbol{t}_{j}, R_{j} \}$ in that:

\begin{enumerate}
	\item To obtain $\{ \boldsymbol{t}_{j} \}$ with given $\{ R_{j} \}$, we simply take the minimizer of (\ref{eq-node loss}):
\begin{equation}\label{eq-node vector}
\boldsymbol{t}_j = \argmin_{\boldsymbol{t}} NodeLoss(\boldsymbol{t};\mathcal{I}_{j}),
\end{equation}
where $\mathcal{I}_{j}$ denotes the index set for $R_{j}$.


\item To obtain $\{ R_{j} \}$, we recursively perform binary split until there are $J$-terminal nodes.
\end{enumerate}	
The key to the second point is to explicitly give the node split gain. Suppose an internal node with $n$ training examples ($n=N$ for the root node), we fix on some feature and re-index all the $n$ examples according to their sorted feature values. Now we need to find the index $n'$ with $1 < n' < n$ that maximizes the node gain defined as loss reduction after a division between the $n'$-th and $(n'+1)$-th examples:
\begin{equation}\label{eq-node gain}
\begin{split}
    &NodeGain(n') = NodeLoss(\boldsymbol{t}^*;\mathcal{I}) - \\
    &(NodeLoss(\boldsymbol{t}_L^*;\mathcal{I}_L) + NodeLoss(\boldsymbol{t}_R^*; \mathcal{I}_R))
\end{split}
\end{equation}
where $\mathcal{I}=\{1,\ldots,n\}$, $\mathcal{I}_L=\{1,\ldots,n'\}$ and $\mathcal{I}_R=\{n'+1,\ldots,n\}$; $\boldsymbol{t}^*$, $\boldsymbol{t}_L^*$ and $\boldsymbol{t}_R^*$ are the minimizers of (\ref{eq-node loss}) with index sets $\mathcal{I}$, $\mathcal{I}_L$ and $\mathcal{I}_R$, respectively. Generally, this searching applies to all features. The best division resulting to largest (\ref{eq-node gain}) is then recorded to perform the actual node split.

Note that (\ref{eq-node gain}) arises in the context of an $O(N\times D)$ outer loop, where $D$ is number of features. However, a na\"{i}ve summing of the losses for (\ref{eq-node loss}) incurs an additional $O(N)$ factor in complexity, which finally results in an unacceptable $O(N^{2}D)$ complexity for a single boosting iteration.

A workaround is to use a Newton descent method for which both the gradient and Hessian can be incrementally computed. Let $\gv$, $\Hm$ respectively be the $K \times 1$ gradient vector and $K \times K$ Hessian matrix at $\tv=\zerov$. By dropping the constant $NodeLoss(\zerov;\Is)$ that is irrelevant to $\tv$, the Taylor expansion of (\ref{eq-node loss}) w.r.t. $\tv$ up to second order is:
\begin{equation}\label{eq-node loss Taylor}
\begin{split}
    loss(\boldsymbol{t};\mathcal{I})  = \boldsymbol{g}^{T}\boldsymbol{t} + \frac{1}{2} \boldsymbol{t}^{T} \boldsymbol{H} \boldsymbol{t} \\
    t_{1}+\ldots+t_{K} = 0, \quad \tv \in \RR^K
\end{split}
\end{equation}
By noting the additive separability of (\ref{eq-node loss Taylor}) and using some matrix derivatives, we have
\begin{tabular}{cc}
    \parbox{0.2\textwidth} {\begin{equation} \label{eq-Taylor d1}
        \boldsymbol{g} = -\sum_{i \in I} \gv_i
    \end{equation} } &
    \parbox{0.23\textwidth} { \begin{equation}\label{eq-Taylor d2}
        \boldsymbol{H} = \sum_{i \in I} \Hm_i
    \end{equation} } \\

    \parbox{0.2\textwidth} { \begin{equation}\label{eq-Taylor d1_sample}
      \gv_i =  \boldsymbol{r}_{i} - \boldsymbol{p}_{i}
    \end{equation} } &
    \parbox{0.23\textwidth} { \begin{equation}\label{eq-Taylor-d2-sample}
      \Hm_i = \Phat - \boldsymbol{p}_{i} \boldsymbol{p}_{i}^{T}
    \end{equation} } \\
\end{tabular}
where the diagonal matrix $\Phat=diag(p_{i1},\ldots,p_{iK})$. We then use (\ref{eq-node loss Taylor}) to compute the approximated node loss in (\ref{eq-node gain}). Thanks to the additive form, both (\ref{eq-Taylor d1}) and (\ref{eq-Taylor d2}) can be incrementally/decrementally computed in constant time when the split searching proceeds from one training example to the next. Therefore, the computation of (\ref{eq-node loss Taylor}) eliminates the $O(N)$ complexity in the na\"{i}ve summing of losses.\footnote{In Real AdaBoost.MH, such a second order approximation is not necessary (although possible, cf. \cite{Zou-Multi}). Due to the special form of the exponential loss and the absence of a sum-to-zero constraint, there exists analytical solution for the node loss (\ref{eq-node loss}) by simply setting the derivative to $\boldsymbol{0}$. Here also, the computation can be incremental/decremental. Since the loss design and AdaBoost.MH are not our main interests, we do not discuss this further.}

\subsection{Properties of Approximated Node Loss}\label{sec-loss-prop}

To minimise (\ref{eq-node loss Taylor}), we give some properties for (\ref{eq-node loss Taylor}) that should be taken into account when seeking a solver. We begin with the sum-to-zero constraint. The probability estimate $p_k$ in the Logit loss (\ref{eq5-log loss}) must be non-zero and sum-to-one, which is ensured by the link function (\ref{eq-link}). Such a link, in turn, means that $p_k$ is unchanged by adding an arbitrary constant to each component in $\Fv$. As a result, the single example loss (\ref{eq5-log loss}) is invariant to moving it along an all-1 vector $\onev$. That is,
\begin{equation}\label{eq-loss-inv}
  L(y_i, \Fv_i + c\onev) = L(y_i, \Fv_i),
\end{equation}
where $c$ is an arbitrary real constant (Note that $\onev$ is, coincidentally, the orthogonal complement to the space defined by sum-to-zero constraint). This property also carries over to the approximated node loss (\ref{eq-node loss Taylor}):
\begin{theo-stzH}\label{theo-inv}
  $loss(\tv;\Is) = loss(\tv+c\onev;\Is)$.
\end{theo-stzH}
This is obvious by noting the additive separability in (\ref{eq-node loss Taylor}), as well as that $\gv_i^T\onev = 0$, $\onev^T\Hm_i\onev = 0$ holds since $\pv_i$ is sum-to-one.

For the Hessian, we have $rank(\Hm) \leq rank(\Hm_i)$ by noting the additive form in (\ref{eq-Taylor d1}). In \cite{Li-ABCLogit} it is shown that $det\Hm_i = 0$ by brute-force determinant expansion. Here we give a stronger property:
\begin{theo-stzH}\label{theo-H}
  $\Hm_i$ is a positive semi-definite matrix such that 1) $rank(\Hm_i) = \kappa-1$, where $\kappa$ is the number of non-zero elements in $\pv_i$; 2) $\onev$ is the eigenvector for eigenvalue 0.
\end{theo-stzH}
The proof can be found in this paper's extended version \cite{this-ext}.

The properties shown above indicate that 1) $\Hm$ is singular, so that unconstrained Newton descent is not applicable here, and 2) $rank(H)$ could be as high as $K-1$, which prohibits the application of the standard fast quadratic solver designed for low rank Hessians. In the following we propose to address this problem via block coordinate descent, a technique that has been successfully used in training SVMs~\cite{Bottou-solver}.




\subsection{Block Coordinate Descent}\label{sec-block coord}
For the variable $\boldsymbol{t}$ in (\ref{eq-node loss Taylor}), we only choose two (the least possible number due to the sum-to-zero constraint) coordinates, \ie a class pair, to update while keeping the others fixed. Suppose we have chosen the $r$-th and the $s$-th coordinate (how to do so is deferred to next subsection). Let $t_{r} = t$ and $t_{s} = -t$ be the free variables (such that $t_r + t_s = 0$) and $t_k = 0$ for $k \neq r$, $k \neq s$. Plugging these into (\ref{eq-node loss Taylor}) yields an unconstrained one dimensional quadratic problem with regards to the scalar variable $t$:
\begin{equation}\label{eq-node loss Taylor scalar}
    loss(t) = g^{T}t + \frac{1}{2} h t^{2}
\end{equation}
where the gradient and Hessian collapse to scalars:
\begin{equation}\label{eq-Taylor d1 scalar}
    g = -\sum_{i \in I} \left(
    (r_{i,r} - p_{i,r}) - (r_{i,s} - p_{i,s})
    \right)
\end{equation}
\begin{equation}\label{eq-Taylor d2 scalar}
    h = \sum_{i \in I} \left(
    p_{i,r}(1-p_{i,r}) + p_{i,s}(1-p_{i,s}) + 2p_{i,r}p_{i,s}
    \right),
\end{equation}
To this extent, we are able to obtain the analytical expression for the minimizer and minimum of (\ref{eq-node loss Taylor scalar}):
\begin{equation}\label{eq-newton step 1d}
    t^{\ast} = \argmin_{t} loss(t) = -\frac{g}{h}
\end{equation}
\begin{equation}\label{eq-newton dec 1d}
    loss(t^{\ast}) = -\frac{g^{2}}{2h}
\end{equation}
by noting the non-negativity of (\ref{eq-Taylor d2 scalar}).

Based on (\ref{eq-newton step 1d}), node vector (\ref{eq-node vector}) can be approximated as
\begin{equation}\label{eq-node vector appro}
\boldsymbol{t}_{mj} =
\begin{cases}
    +(-g/h) \quad k = r \\
    -(-g/h) \quad k = s \\
    0       \quad\quad\quad\quad otherwise
\end{cases}
\end{equation}
where $g$ and $h$ are respectively computed by using (\ref{eq-Taylor d1 scalar}) and (\ref{eq-Taylor d2 scalar}) with index set $\mathcal{I}_{mj}$. Based on (\ref{eq-newton dec 1d}), the node gain (\ref{eq-node gain}) can be approximated as
\begin{equation} \label{eq-node gain appro}
    NodeGain(n') = \frac{g_L^{2}}{2h_L} + \frac{g_R^{2}}{2h_R} - \frac{g^{2}}{2h},
\end{equation}
where $g$ (or $g_L$, $g_R$) and $h$ (or $h_L$, $h_R$) are computed by using (\ref{eq-Taylor d1 scalar}) and (\ref{eq-Taylor d2 scalar}) with index set $\mathcal{I}$ (or $\mathcal{I}_L$, $\mathcal{I}_R$).

\subsection{Class Pair Selection}\label{sec-subset sel}
In~\cite{Bottou-solver} two methods for selecting $(r,s)$ are proposed. One is based on a first order approximation. Let $t_r$ and $t_s$ be the free variables and the rest be fixed to $0$. For a $\boldsymbol{t}$ with sufficiently small fixed length, let $t_{r}=\epsilon$ and $t_{s}=-\epsilon$ where $\epsilon>0$ is some small enough constant. The first order approximation of (\ref{eq-node loss Taylor}) is:
\begin{equation}\label{eq20-1 order exp}
    loss(\boldsymbol{t}) \approx loss(\boldsymbol{0}) + \boldsymbol{g}^{T}\boldsymbol{t} = loss(\boldsymbol{0}) - \epsilon(-g_r - (-g_s))
\end{equation}
It follows that the indices $r$, $s$ resulting in largest decrement to (\ref{eq20-1 order exp}) are:
\begin{equation}\label{eq-subsel 1}
\begin{split}
    &r = \argmax_{k} \left \{ -g_{k} \right \} \\
    &s = \argmin_{k} \left \{ -g_{k} \right \}.
\end{split}
\end{equation}
Another method that can be derived in a similar way takes into account the second order information:
\begin{equation}\label{eq-subsel 2}
\begin{split}
    &r = \argmax_{k} \left \{ -g_{k} \right \} \\
    &s = \argmax_{k} \left \{ \frac {(g_{r}-g_{k})^{2}} {h_{rr}+h_{kk}-2h_{rk}} \right \},
\end{split}
\end{equation}
Both methods are $O(K)$ procedures that are better than the $K\times(K-1)/2$ na\"{i}ve enumeration. However, in our implementation we find that (\ref{eq-subsel 2}) achieves better results for AOSO-LogitBoost.


Pseudocode for AOSO-LogitBoost is given in Algorithm~\ref{alg-AOSOBoost}.

\begin{algorithm}
\caption{AOSO-LogitBoost. $v$ is shrinkage factor that controls learning rate.}\label{alg-AOSOBoost}
\begin{algorithmic}[1]
\STATE $F_{ik}=0$, \quad $k=1,\ldots,K$, $i=1,\ldots,N$
\FOR{$m = 1$ to $M$}
    \STATE $p_{i,k}=\frac {\exp(F_{i,k})} {\sum_{j=1}^K\exp(F_{i,j})}$, $k=1,\ldots,K$, $i=1,\ldots,N$.
    \STATE Obtain $\{ R_{mj} \}_{j=1}^J$ by recursive region partition. Node split gain is computed as (\ref{eq-node gain appro}), where the class pair ($r$, $s$) is selected using (\ref{eq-subsel 2}) .
    \STATE Compute $ \{ \boldsymbol{t}_{mj} \}_{j=1}^J $ by (\ref{eq-node vector appro}), where the class pair ($r$, $s$) is selected using (\ref{eq-subsel 2}) .
    \STATE $\boldsymbol{F_i} = \boldsymbol{F_i} + v\sum_{j=1}^J \boldsymbol{t}_{mj} I(\boldsymbol{x}_i \in R_{mj})$, \quad $i=1,\ldots,N$.
\ENDFOR
\end{algorithmic}
\end{algorithm}

\section{Comparison to (ABC-)LogitBoost}\label{sec-comp}

In this section we compare the derivations of LogitBoost and ABC-LogitBoost and provide some intuition for observed behaviours in the experiments in Section~\ref{sec-exp}.

\subsection{ABC-LogitBoost}\label{sec-compABC}
To solve (\ref{eq-fm}) with a sum-to-zero constraint, ABC-LogitBoost uses $K-1$ independent trees:
\begin{equation}
f_k =
\begin{cases}
    \sum_j t_{jk}I(x \in R_{jk}) \quad k \neq b\\
    -\sum_{l \neq b} f_l \quad\quad\quad\quad k = b.
\end{cases}
\end{equation}
In~\cite{Li-ABCLogit}, the so-called base class $b$ is selected by exhaustive search per iteration, \ie trying all possible $b$, which involves growing $K(K-1)$ trees. To reduce the time complexity, Li also proposed other methods. In~\cite{Li-FastABC}, $b$ is selected only every several iterations, while in~\cite{Li-ABC}, $b$ is, intuitively, set to be the class that leads to largest loss reduction at last iteration.


In ABC-LogitBoost the sum-to-zero constraint is explicitly considered when deriving the node value and the node split gain for the scalar regression tree. Indeed, they are the same as (\ref{eq-node vector appro}) and (\ref{eq-node gain appro}) in this paper, although derived using a slightly different motivation. In this sense, ABC-LogitBoost can be seen as a special form of the AOSO-LogitBoost since: 1) For each tree, the class pair is fixed for every node in ABC, while it is selected adaptively in AOSO, and 2) $K-1$ trees are added per iteration in ABC (using the same set of probability estimates $\{p_i\}_{i=1}^{N}$), while only one tree is added per iteration by AOSO (and $\{p_i\}_{i=1}^{N}$ are updated as soon as each tree is added).

Since two changes are made to ABC-LogitBoost, an immediate question is what happens if we only make one? That is, what happens if one vector tree is added per iteration for a single class pair selected only for the root node and shared by all other tree nodes, as in ABC, but the $\{p_i\}_{i=1}^{N}$ are updated as soon as a tree is added, as in AOSO. This was tried but unfortunately, \textbf{degraded performance} was observed for this combination so the results are not reported here.

From the above analysis, we believe the more flexible model (as well as the model updating strategy) in AOSO is what contributes to its improvement over ABC, as seen section \ref{sec-exp}).

\subsection{LogitBoost}

In the original LogitBoost~\cite{Friedman-additive}, the Hessian matrix (\ref{eq-Taylor-d2-sample}) is approximated diagonally. In this way, the $\boldsymbol{f}$ in (\ref{eq-fm}) is expressed by $K$ uncoupled scalar tress:
\begin{equation}
    f_k = \sum_j t_{jk}I(x \in R_{jk}), \quad k=1,2,\ldots,K \\
\end{equation}
with the gradient and Hessian for computing node value and node split gain given by:
\begin{equation}
    g_k = -\sum_{i \in \mathcal{I}} (r_{i,k} - p_{i,k}), \quad
    h_k = -\sum_{i \in \mathcal{I}} p_{i,k}(1-p_{i,k}).
\end{equation}
Here we use the subscript $k$ for $g$ and $h$ to emphasize the $k$-th tree is built independently to the other $K-1$ trees (\ie the sum-to-zero constraint is dropped). Although this simplifies the mathematics, such an aggressive approximation turns out to harm both classification accuracy and convergence rate, as shown in Li's experiments~\cite{Li-ABCLogitExp}.
\begin{table}
\caption{Datasets used in our experiments.} \label{tab-datasets}
\begin{center}
\tiny{
\begin{tabular}{l llll }
\hline\hline
datasets & K & $\#$features & $\#$training & $\#$test\\
\hline
Poker525k &10 &25 &525010 &500000 \\
Poker275k &10 &25 &275010 &500000 \\
Poker150k &10 &25 &150010 &500000 \\
Poker100k &10 &25 &100010 &500000 \\
Poker25kT1 &10 &25 &25010 &500000 \\
Poker25kT2 &10 &25 &25010 &500000 \\
Covertype290k &7 &54 &290506 &290506 \\
Covertype145k &7 &54 &145253 &290506 \\
Letter & 26 & 16 & 16000 & 4000 \\
Letter15k & 26 & 16 & 15000 & 5000 \\
Letter2k & 26 & 16 & 2000 & 18000 \\
Letter4K & 26 & 16 & 4000 & 16000\\
Pendigits & 10 & 16 & 7494 & 3498 \\
Zipcode & 10 & 256 & 7291 & 2007\\
(a.k.a. USPS) & & & &\\
Isolet & 26 & 617 & 6238 & 1559\\
Optdigits & 10 & 64 & 3823 & 1797\\
\hline
Mnist10k & 10 & 784 & 10000 & 60000 \\
M-Basic & 10 & 784 & 12000 & 50000 \\
M-Image & 10 & 784 & 12000 & 50000 \\
M-Rand & 10 & 784 & 12000 & 50000 \\
M-Noise1 & 10 & 784 & 10000 & 2000 \\
M-Noise2 & 10 & 784 & 10000 & 2000 \\
M-Noise3 & 10 & 784 & 10000 & 2000 \\
M-Noise4 & 10 & 784 & 10000 & 2000 \\
M-Noise5 & 10 & 784 & 10000 & 2000 \\
M-Noise6 & 10 & 784 & 10000 & 2000 \\
\hline\hline
\end{tabular} 
} 
\end{center}
\end{table}

\section{Experiments}\label{sec-exp}
In this section we compare AOSO-LogitBoost with ABC-LogitBoost, which was shown to outperform original LogitBoost in Li's experiments~\cite{Li-ABCLogit,Li-ABCLogitExp}. We test AOSO on all the datasets used in \cite{Li-ABCLogit,Li-ABCLogitExp}, as listed in Table~\ref{tab-datasets}. In the top section are UCI datasets and in the bottom are Mnist datasets with many variations (see \cite{Li-EmpExp} for detailed descriptions).\footnote{Code and data are available at http://ivg.au.tsinghua.edu.cn/index.php?n=People.PengSun}
To exhaust the learning ability of (ABC-)LogitBoost, Li let the boosting stop when either the training converges (\ie the loss (\ref{eq-total loss}) approaches $0$, implemented as $\leq 10^{-16}$) or a maximum number of iterations, $M$, is reached. Test errors at last iteration are simply reported since no obvious over-fitting is observed. By default, $M=10000$, while for those large datasets (\emph{\textbf{Covertype290k, Poker525k, Pokder275k, Poker150k, Poker100k}}) $M=5000$~\cite{Li-ABCLogit,Li-ABCLogitExp}. We adopt the same criteria, except that our maximum iterations $M_{AOSO} = (K-1) \times M_{ABC}$, where $K$ is the number of classes. Note that only one tree is added at each iteration in AOSO, while $K-1$ are added in ABC. Thus, this correction compares the same maximum number of trees for both AOSO and ABC.
\begin{table}
\caption{Test classification errors on \emph{\textbf{Mnist10k}}. In each $J$-$v$ entry, the first entry is for ABC-LogitBoost and the second for AOSO-LogitBoost. Lower one is in bold.} \label{tab-mnist10k}
\begin{center}
\tiny{
\begin{tabular}{l llll}
\hline\hline
&$v=0.04$ &$v=0.06$ &$v=0.08$ &$v=0.1$\\
\hline
$J=4$  &2630 \textbf{2515} &2600 \textbf{2414} &2535 \textbf{2414} &2522 \textbf{2392} \\
$J=6$  &2263 \textbf{2133} &2252 \textbf{2146} &2226 \textbf{2146} &2223 \textbf{2134} \\
$J=8$  &2159 \textbf{2055} &2138 \textbf{2046} &2120 \textbf{2046} &2143 \textbf{2055} \\
$J=10$ &2122 \textbf{2010} &2118 \textbf{1980} &2091 \textbf{1980} &2097 \textbf{2014} \\
$J=12$ &2084 \textbf{1968} &2090 \textbf{1965} &2090 \textbf{1965} &2095 \textbf{1995} \\
$J=14$ &2083 \textbf{1945} &2094 \textbf{1938} &2063 \textbf{1938} &2050 \textbf{1935} \\
$J=16$ &2111 \textbf{1941} &2114 \textbf{1928} &2097 \textbf{1928} &2082 \textbf{1966} \\
$J=18$ &2088 \textbf{1925} &2087 \textbf{1916} &2088 \textbf{1916} &2097 \textbf{1920} \\
$J=20$ &2128 \textbf{1930} &2112 \textbf{1917} &2095 \textbf{1917} &2102 \textbf{1948} \\
$J=24$ &2174 \textbf{1901} &2147 \textbf{1920} &2129 \textbf{1920} &2138 \textbf{1903} \\
$J=30$ &2235 \textbf{1887} &2237 \textbf{1885} &2221 \textbf{1885} &2177 \textbf{1917} \\
$J=40$ &2310 \textbf{1923} &2284 \textbf{1890} &2257 \textbf{1890} &2260 \textbf{1912} \\
$J=50$ &2353 \textbf{1958} &2359 \textbf{1910} &2332 \textbf{1910} &2341 \textbf{1934} \\
\hline\hline
\end{tabular} 
}
\end{center}
\end{table}

The most important tuning parameters in LogitBoost are the number of terminal nodes $J$, and the shrinkage factor $v$. In \cite{Li-ABCLogit,Li-ABCLogitExp}, Li reported results of (ABC-)LogitBoost for a number of $J$-$v$ combinations. We report the corresponding results for AOSO-LogitBoost for the same combinations. In the following, we intend to show that \textbf{for nearly all $J$-$v$ combinations, AOSO-LogitBoost has lower classification error and faster convergence rates than ABC-LogitBoost}.

\subsection{Classification Errors}\label{sec-clserr}
Table~\ref{tab-mnist10k} shows results of various $J$-$v$ combinations for a representative datasets. Results on more datasets can be found in this paper's extended version \cite{this-ext}.

\begin{table*}
\caption{Summary of test classification errors. Lower one is in bold. Middle panel: $J=20$,$v=0.1$ except for \emph{\textbf{Poker25kT1}} and \emph{\textbf{Poker25kT2}} on which $J$, $v$ are chosen by validation (See the text in \ref{sec-clserr}); Right panel: the overall best. Dash "-" means unavailable in \cite{Li-ABCLogit}\cite{Li-ABCLogitExp}. Relative improvements ($R$) and $P$-values ($pv$) are given.} \label{tab-summary}
\begin{center}
\tiny{
\begin{tabular}{ll | llll | llll  }
\hline\hline
Datasets &$\#$tests &ABC &AOSO &$R$ &$pv$ &ABC$^*$ &AOSO$^*$ &$R$ &$pv$\\
\hline
Poker525k     &500000  &1736         &\textbf{1537 } &0.1146	 &0.0002  &-            &-               &-       &-      \\
Poker275k     &500000  &2727         &\textbf{2624 } &0.0378	 &0.0790  &-            &-               &-       &-      \\
Poker150k     &500000  &5104         &\textbf{3951 } &0.2259	 &0.0000  &-            &-               &-       &-      \\
Poker100k     &500000  &13707        &\textbf{7558 } &0.4486	 &0.0000  &-            &-               &-       &-      \\
Poker25kT1    &500000  &37345        &\textbf{31399} &0.1592	 &0.0000  &37345        &\textbf{31399}  &0.1592  &0.0000 \\
Poker25kT2    &500000  &36731        &\textbf{31645} &0.1385	 &0.0000  &36731        &\textbf{31645}  &0.1385  &0.0000 \\
Covertype290k &290506  &9727         &\textbf{9586 } &0.0145	 &0.1511  &-            &-               &-       &-      \\
Covertype145k &290506  &13986        &\textbf{13712} &0.0196	 &0.0458  &-            &-               &-       &-      \\
Letter        &4000    &\textbf{89}  &92             &-0.0337	 &0.5892  &89           &\textbf{88}     &0.0112  &0.4697 \\
Letter15k     &5000    &\textbf{109} &116            &-0.0642	 &0.6815  &-            &-               &-       &-      \\
Letter4k      &16000   &1055         &\textbf{991  } &0.0607	 &0.0718  &1034         &\textbf{961}    &0.0706  &0.0457 \\
Letter2k      &18000   &2034         &\textbf{1862 } &0.0846	 &0.0018  &1991         &\textbf{1851}   &0.0703  &0.0084 \\
Pendigits     &3498    &100          &\textbf{83   } &0.1700	 &0.1014  &90           &\textbf{81}     &0.1000  &0.2430 \\
Zipcode       &2007    &\textbf{96}  &99             &-0.0313	 &0.5872  &\textbf{92}  &94              &-0.0217 &0.5597 \\
Isolet        &1559    &65           &\textbf{55   } &0.1538	 &0.1759  &55           &\textbf{50}     &0.0909  &0.3039 \\
Optdigits     &1797    &55           &\textbf{38   } &0.3091	 &0.0370  &38           &\textbf{34}     &0.1053  &0.3170 \\
Mnist10k      &60000   &2102         &\textbf{1948 } &0.0733	 &0.0069  &2050         &\textbf{1885}   &0.0805  &0.0037 \\
M-Basic       &50000   &1602         &\textbf{1434 } &0.1049	 &0.0010  &-            &-               &-       &-      \\
M-Rotate      &50000   &5959         &\textbf{5729 } &0.0386	 &0.0118  &-            &-               &-       &-      \\
M-Image       &50000   &4268         &\textbf{4167 } &0.0237	 &0.1252  &4214         &\textbf{4002}   &0.0503  &0.0073 \\
M-Rand        &50000   &4725         &\textbf{4588 } &0.0290	 &0.0680  &-            &-               &-       &-      \\
M-Noise1      &2000    &234          &\textbf{228  } &0.0256	 &0.3833  &-            &-               &-       &-      \\
M-Noise2      &2000    &237          &\textbf{233  } &0.0169	 &0.4221  &-            &-               &-       &-      \\
M-Noise3      &2000    &238          &\textbf{233  } &0.0210	 &0.4031  &-            &-               &-       &-      \\
M-Noise4      &2000    &238          &\textbf{233  } &0.0210	 &0.4031  &-            &-               &-       &-      \\
M-Noise5      &2000    &227          &\textbf{214  } &0.0573	 &0.2558  &-            &-               &-       &-      \\
M-Noise6      &2000    &201          &\textbf{191  } &0.0498	 &0.2974  &-            &-               &-       &-      \\
\hline\hline
\end{tabular} 
}
\end{center}
\end{table*}
In Table~\ref{tab-summary} we summarize the results for all datasets. In \cite{Li-ABCLogit}, Li reported that ABC-LogitBoost is insensitive to $J$, $v$ on all datasets except for \emph{\textbf{Poker25kT1}} and \emph{\textbf{Poker25kT2}}. Therefore, Li summarized classification errors for ABC simply with $J=20$ and $v=0.1$, except that on \emph{\textbf{Poker25kT1}} and \emph{\textbf{Poker25kT2}} errors are reported by using the other's test set as a validation set. Based on the same criteria we summarize AOSO in the middle panel of Table~\ref{tab-summary} where the test errors as well as the improvement relative to ABC are given. In the right panel of Table~\ref{tab-summary} we provide the comparison for the best results achieved over all $J$-$v$ combinations when the corresponding results for ABC are available in \cite{Li-ABCLogit} or \cite{Li-ABCLogitExp}.
\begin{table}
\caption{$\#$trees added when convergence on selected datasets. $R$ stands for the ratio AOSO/ABC.} \label{tab-conv}
\begin{center}
\tiny{
\begin{tabular}{l lll }
\hline\hline
\quad &Mnist10k &M-Rand &M-Image   \\
\hline
ABC  &7092   &15255   &14958         \\
$R$  &0.7689 &0.7763  &0.8101    \\
\hline\hline
\quad &Letter15k &Letter4k & Letter2k   \\
\hline
ABC &45000 &20900   &13275 \\
$R$ &0.5512 &0.5587  &0.5424  \\
\hline\hline
\end{tabular} 
}
\end{center}
\end{table}
\begin{table}
\caption{$\#$trees added when convergence on \textbf{\emph{Mnist10k}} for a number of $J$-$v$ combinations. For each $J$-$v$ entry, the first number is for ABC, the second for the ratio AOSO/ABC.} \label{tab-conv-minist10k}
\begin{center}
\tiny{
\begin{tabular}{l lll }
\hline\hline
\quad &$v=0.04$ &$v=0.06$ &$v=0.1$ \\
\hline
$J=4$  &90000  1.0    &90000  1.0    &90000  1.0     \\
$J=6$  &90000  0.7740 &63531  0.7249 &38223  0.7175  \\
$J=8$  &55989  0.7962 &38223  0.7788 &22482  0.7915  \\
$J=10$ &39780  0.8103 &27135  0.7973 &16227  0.8000  \\
$J=12$ &31653  0.8109 &20997  0.8074 &12501  0.8269  \\
$J=14$ &26694  0.7854 &17397  0.8047 &10449  0.8160  \\
$J=16$ &22671  0.7832 &11704  1.0290 &8910   0.8063  \\
$J=18$ &19602  0.7805 &13104  0.7888 &7803   0.7933  \\
$J=20$ &17910  0.7706 &11970  0.7683 &7092   0.7689  \\
$J=24$ &14895  0.7514 &9999   0.7567 &6012   0.7596  \\
$J=30$ &12168  0.7333 &8028   0.7272 &4761   0.7524  \\
$J=40$ &9846   0.6750 &6498   0.6853 &3870   0.6917  \\
$J=50$ &8505   0.6420 &5571   0.6448 &3348   0.6589  \\
\hline\hline
\end{tabular} 
}
\end{center}
\end{table}

We also tested the statistical significance between AOSO and ABC. We assume the classification error rate is subject to some Binomial distribution. Let $z$ denote the number of errors and $n$ the number of tests, then the estimate of error rate $\hat{p}=z/n$ and its variance is $\hat{p}(1-\hat{p})/n$. Subsequently, we approximate the Binomial distribution by a Gaussian distribution and perform a hypothesis test. The $p$-values are reported in Table~\ref{tab-summary}.

For some problems, we note LogitBoost (both ABC and AOSO) outperforms other state-of-the-art classifier such as SVM or Deep Learning. (\emph{e.g.}, the test error rate on \emph{\textbf{Poker}} is $~40\%$ for SVM and $<10\%$ for both ABC and AOSO (even lower than ABC); on \emph{\textbf{M-Image}} it is $16.15\%$ for DBN-1, $8.54\%$ for ABC and $8.33\%$ for AOSO). See this paper's extended version \cite{this-ext} for details. This shows that the AOSO's improvement over ABC does deserve the efforts.

\subsection{Convergence Rate}\label{sec-conv}
\begin{figure}
\begin{center}
\mbox{
  \includegraphics[width=0.16\textwidth]{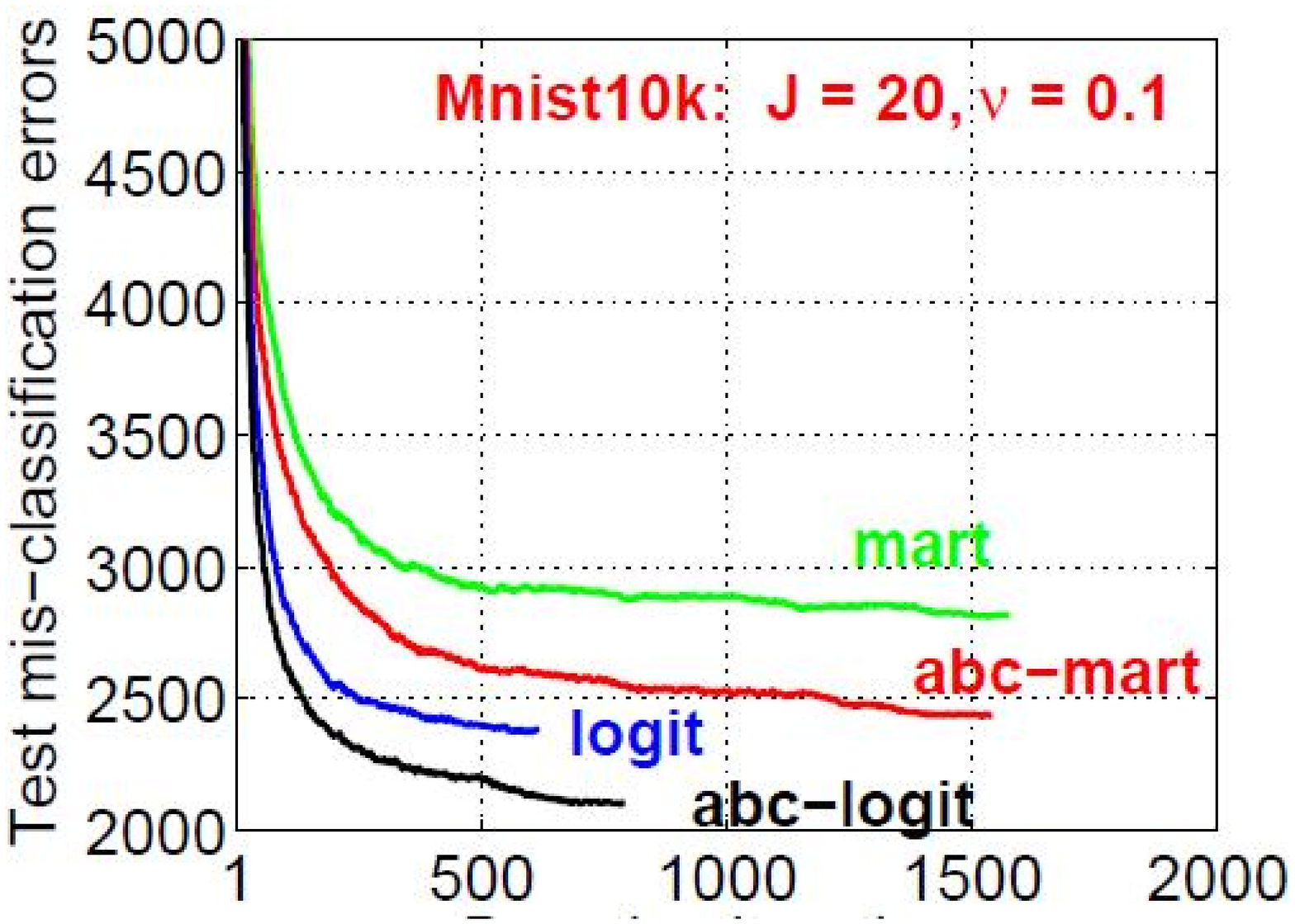}
  \includegraphics[width=0.16\textwidth]{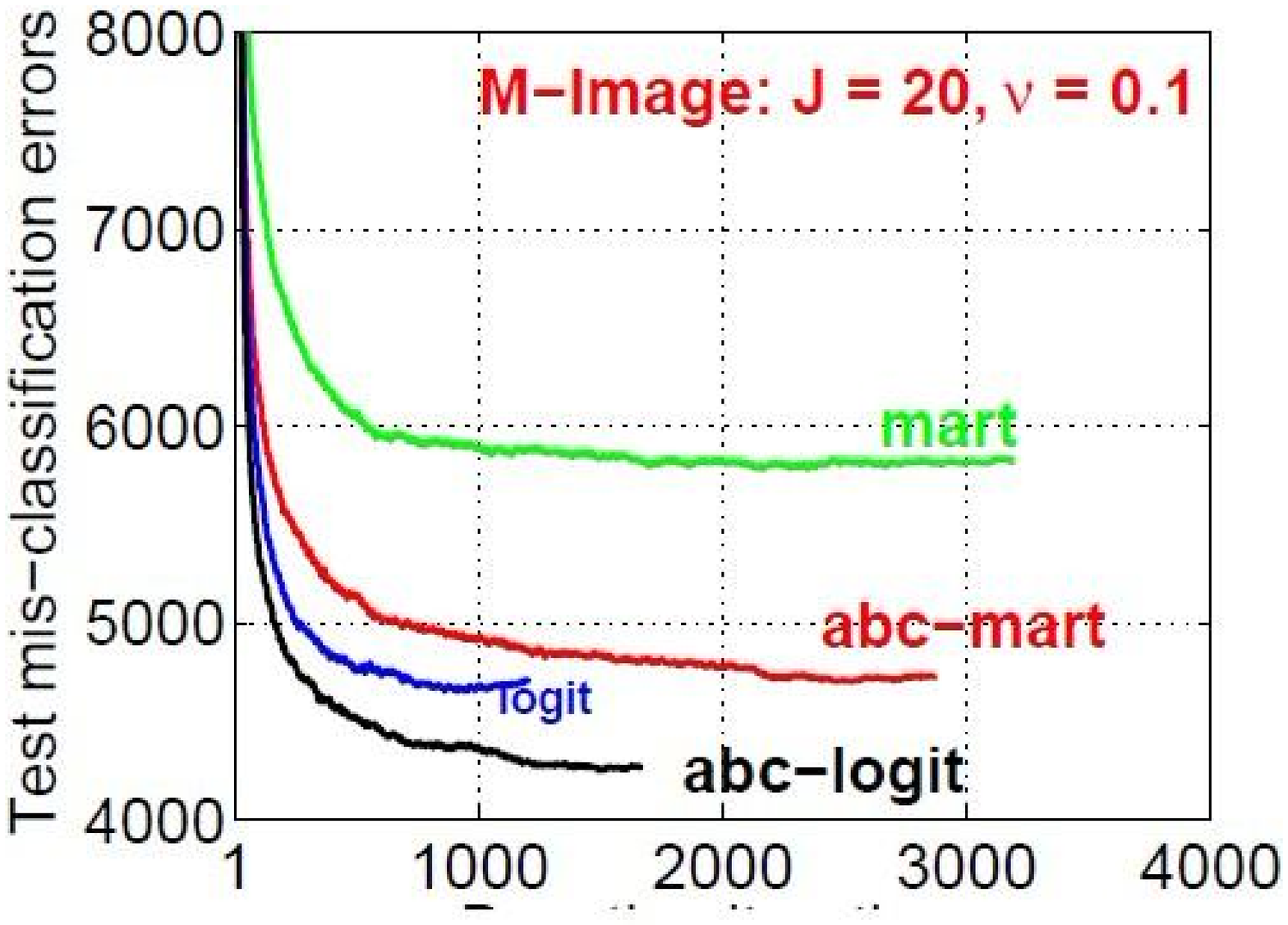}
  \includegraphics[width=0.16\textwidth]{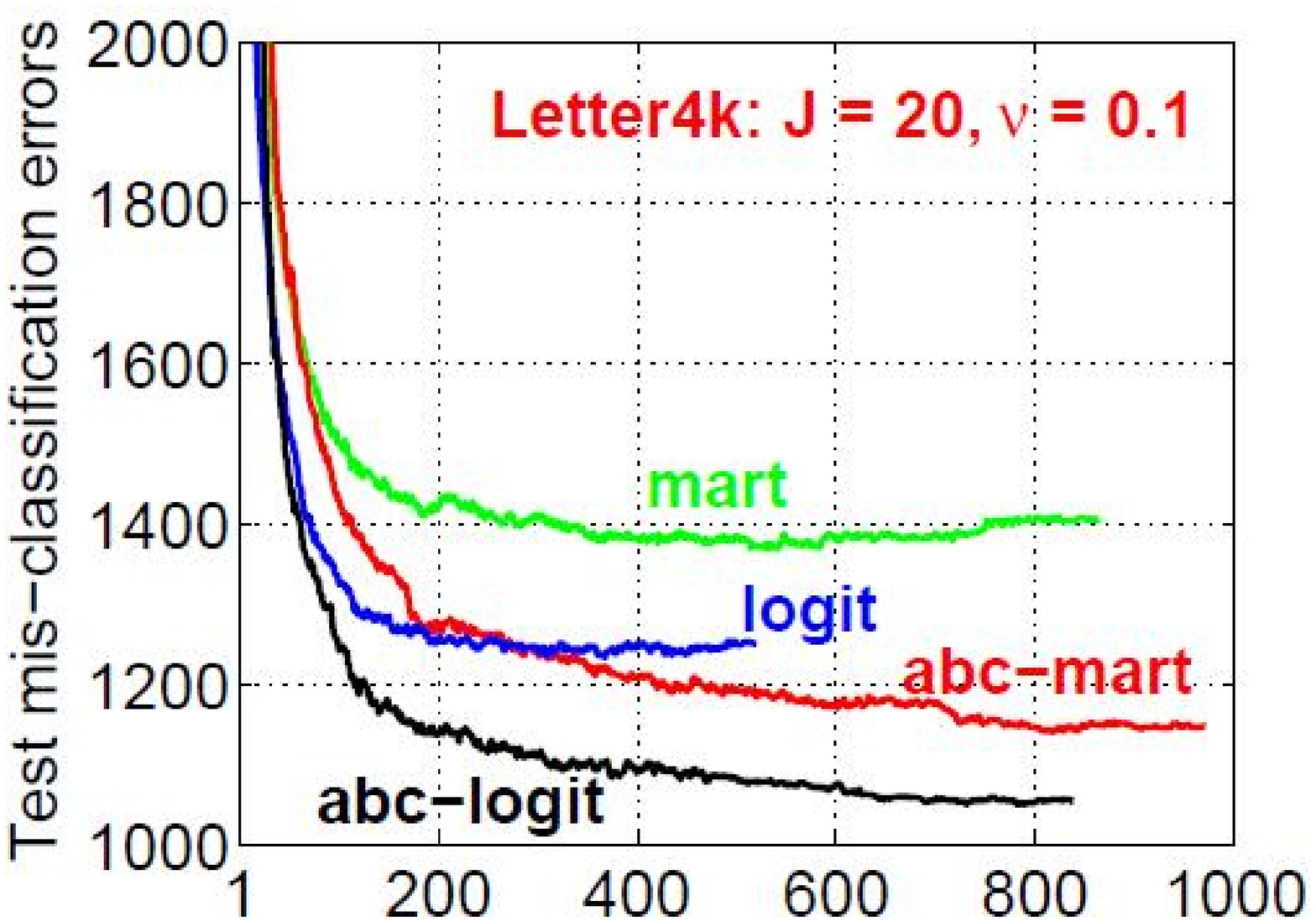}
}
\mbox{
  \includegraphics[width=0.16\textwidth]{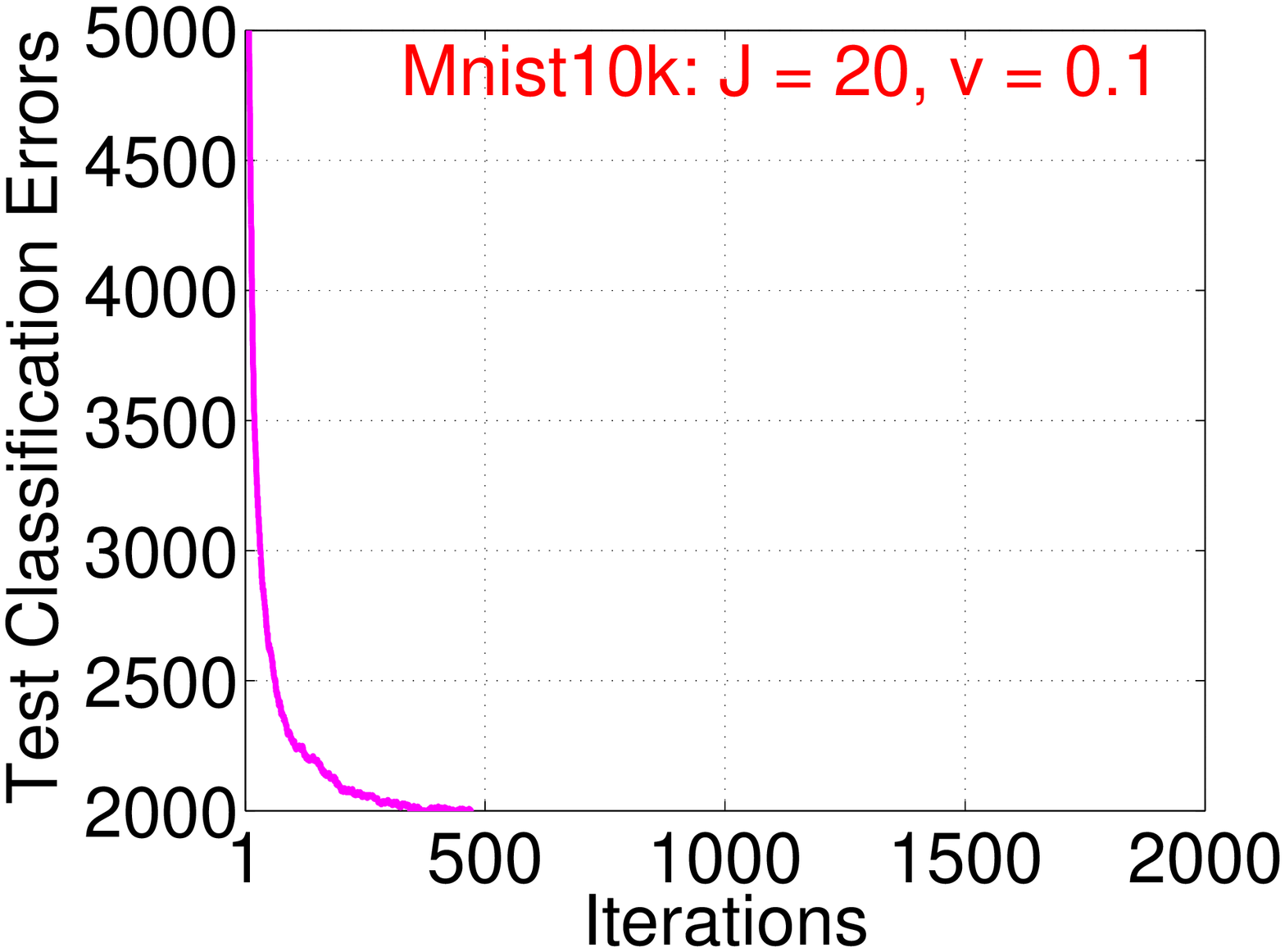}
  \includegraphics[width=0.16\textwidth]{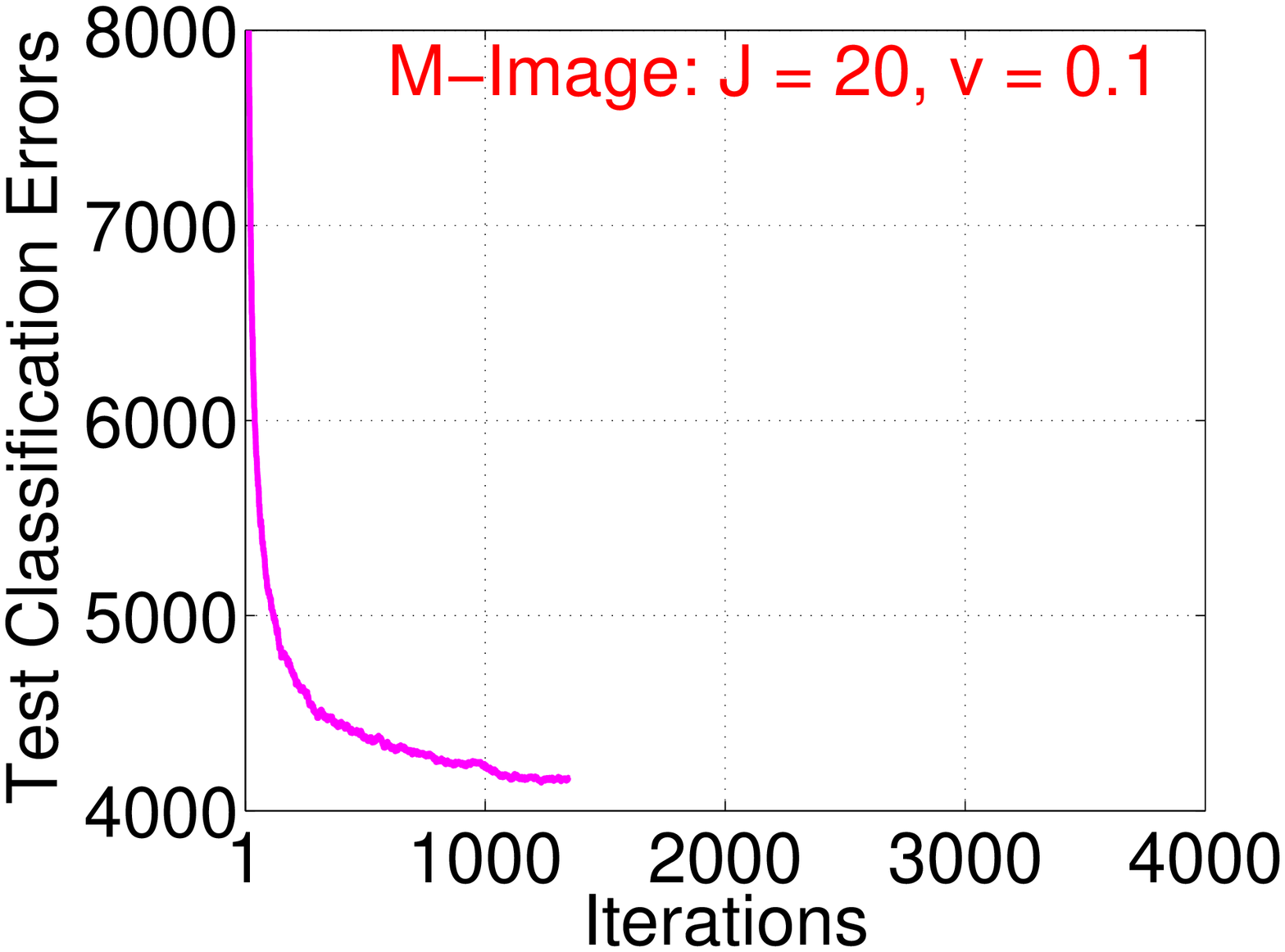}
  \includegraphics[width=0.16\textwidth]{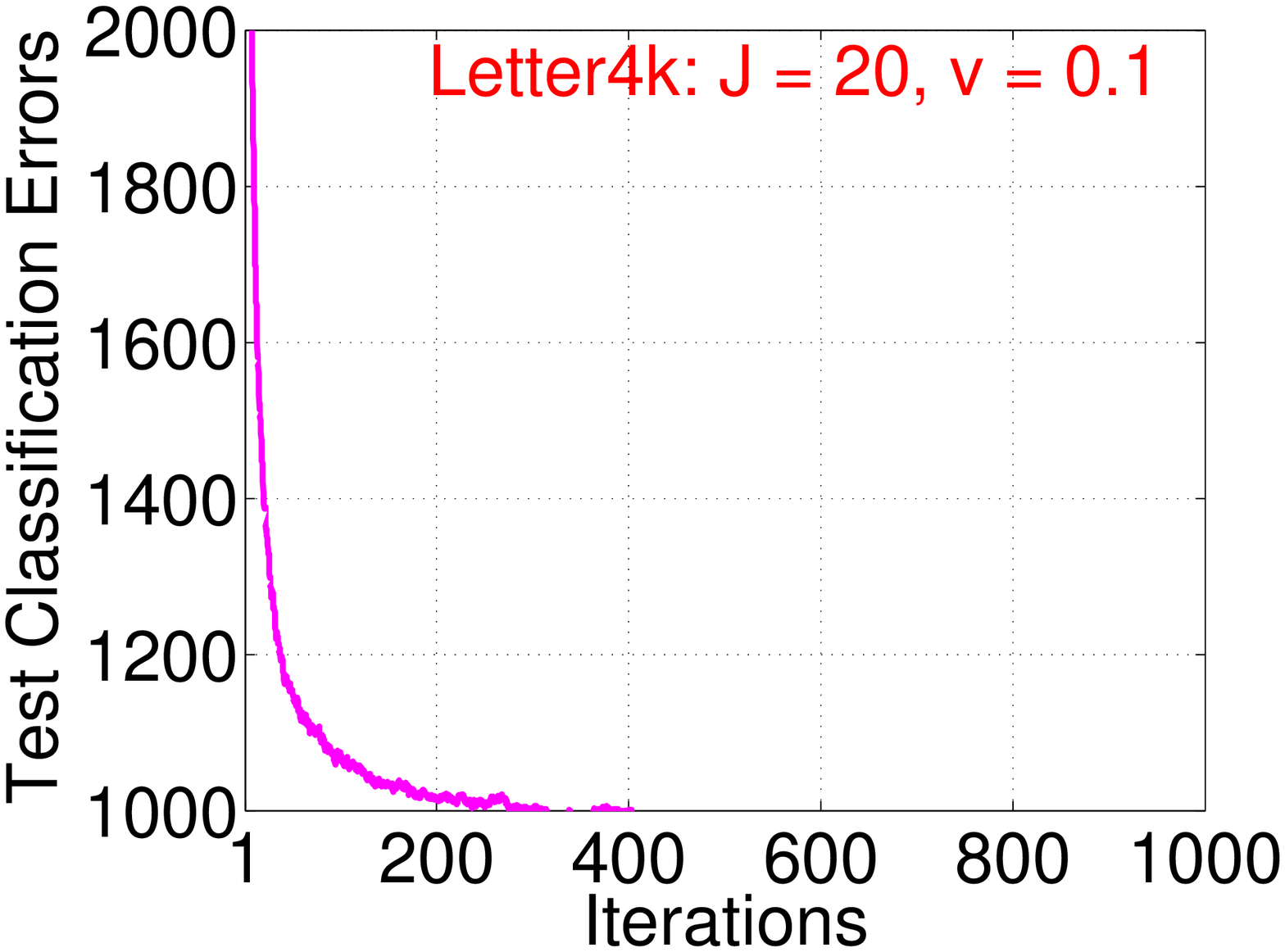}
}
\end{center}
\caption{Errors vs. iterations on selected datasets and parameters. Top row: ABC (copied from \cite{Li-ABCLogit}); Bottom row: AOSO (horizontal axis scaled to compensate the $K-1$ factor).}
\label{fig-other}
\end{figure}
Recall that we stop the boosting procedure if either the maximum number of iterations is reached or it converges (i.e. the loss (\ref{eq-total loss}) $\leq 10^{-16}$). The fewer trees added when boosting stops, the faster the convergence and the lower the time cost for either training or testing. We compare AOSO with ABC in terms of the number of trees added when boosting stops for the results of ABC available in \cite{Li-ABCLogit,Li-ABCLogitExp}. Note that simply comparing number of boosting iterations is unfair to AOSO, since at each iteration only one tree is added in AOSO and $K-1$ in ABC.

Results are shown in Table~\ref{tab-conv} and Table~\ref{tab-conv-minist10k}. Except for when $J$-$v$ is too small, or particularly difficult datasets where both ABC and AOSO reach maximum iterations, we found that trees needed in AOSO are typically only $50\%$ to $80\%$ of those in ABC.

Figure~\ref{fig-other} shows plots for test classification error vs. iterations in both ABC and AOSO and show that  AOSO's test error decreases faster. More plots for AOSO can be found in this paper's extended version \cite{this-ext}.

\section{Conclusions}
We present an improved LogitBoost, namely AOSO-LogitBoost, for multi-class classification. Compared with ABC-LogitBoost, our experiments suggest that our adaptive class pair selection technique results in lower classification error and faster convergence rates.

\subsubsection*{Acknowledgments}
We appreciate Ping Li's inspiring discussion and generous encouragement. Comments from NIPS2011 and ICML2012 anonymous reviewers helped improve the readability of this paper. This work was supported by National Natural Science Foundation of China (61020106004, 61021063, 61005023), The National Key Technology R\&D Program (2009BAH40B03). NICTA is funded by the Australian Government as represented by the Department of Broadband, Communications and the Digital Economy and the ARC through the ICT Centre of Excellence program.

\small{
\bibliography{aosoboost}
}
\bibliographystyle{icml2012}

\end{document}